\documentclass[10pt,twocolumn,letterpaper]{article}

\usepackage{wacv}
\usepackage{times}
\usepackage{epsfig}
\usepackage{graphicx}
\usepackage{amsmath}
\usepackage{amssymb}
\usepackage{color}

\wacvfinalcopy 

\def\wacvPaperID{499} 

\ifwacvfinal\fi
\begin{document}

\title{Multi-scale Convolution Aggregation and Stochastic Feature Reuse for DenseNets}

\author{Mingjie Wang$^{1}$, Jun Zhou$^{2}$, Wendong Mao$^{1}$, Minglun Gong$^{1*}$\\
	~\\
	\small{$^1$Memorial University of Newfoundland, NL, Canada}\\
	\small{$^2$Dalian University of Technology, Dalian, China}}
\maketitle
\ifwacvfinal\thispagestyle{empty}\fi

\begin{abstract}
   Recently, Convolution Neural Networks (CNNs) obtained huge success in numerous vision tasks. In particular, DenseNets have demonstrated that feature reuse via dense skip connections can effectively alleviate the difficulty of training very deep networks {and that reusing features generated by the {initial layers} in all subsequent layers has {strong} impact on performance. To feed even richer information into the network, a novel {adaptive} Multi-scale Convolution Aggregation module is presented in this paper.} Composed of layers for multi-scale convolutions, trainable cross-scale aggregation, maxout, and concatenation, this module is highly non-linear and can boost the accuracy of DenseNet while using much fewer parameters. In addition, due to {high} model complexity, the network with extremely dense feature reuse is prone to overfitting.  {To address this problem,} a regularization method named Stochastic Feature Reuse is also presented.  Through randomly dropping a set of feature maps to be reused for each mini-batch during the training phase, this regularization method reduces training costs and prevents co-adaptation. Experimental results on CIFAR-10, CIFAR-100 and SVHN benchmarks demonstrated the effectiveness of the proposed methods.

\end{abstract}

\section{Introduction}

Recently, deep learning became a dominant field of machine learning for various vision tasks, such as recognition and classification. In particular, Convolutional Neural Networks (CNNs) have achieved an unprecedented success through AlexNet \cite{krizhevsky2012imagenet}, which has incurred a new line of research concentrating on constructing better performing CNNs \cite{szegedy2016rethinking}. {Increasingly deeper architectures are being created and trained based on the observation that, the deeper the network is, the higher-level features it is able to extract}. AlexNets have 5 convolutional layers \cite{krizhevsky2012imagenet}, VGG Nets \cite{simonyan2014very} have 16 or 19, GoogLeNets \cite{szegedy2015going} have 22, and ResNets \cite{he2016deep} feature over 1000 layers employing residual connections. 

As the networks became very deep, two common issues have emerged: gradients explosion and vanishing. To deal with these problems, several creative architectures, such as Highway networks \cite{srivastava2015highway}, Deeply-Supervised Nets \cite{lee2015deeply} and ResNets \cite{he2016deep}, have been designed.  The key ideas are passing information flow from one layer to another via shortcuts or adding ``companion'' objective functions at each hidden layer respectively. Stochastic depth \cite{huang2016deep} trains an ensemble of ResNets with different depth values by randomly dropping a set of layers during the training phase. FractalNets \cite{larsson2016fractalnet} repeatedly utilize a simple expansion rule  to generate an ultra-deep network containing interacting subpaths of different lengths. Based on the above work, DenseNet \cite{huang2017densely} was introduced, which connects each layer to every subsequent layer. As a result, a given layer in DenseNet takes all feature maps extracted by preceding layers as input. This new connection pattern allows DenseNets to obtain significant improvements over the state-of-the-art on several object recognition benchmark tasks.

On another front, inception series \cite{ioffe2015batch,simonyan2014very,szegedy2017inception,szegedy2016rethinking} have been shown to achieve remarkable performance at very low memory costs. This module is composed by convolutions with different kernel size (1$\times$1, 3$\times$3, 5$\times$5) and a 3$\times$3 max pooling, and then concatenates results from the convolutions and pooling. This design strengthens the regularization and scale invariance of extracted features. Recently, feature pyramid networks (FPN) \cite{lin2017feature} and deep layer aggregation \cite{yu2017deep} have been proposed, which aim at exploiting the inherent multi-scale, pyramidal hierarchy of CNNs. Features at different scale levels are merged together to achieve higher accuracy with fewer parameters.

\begin{figure}[t]
	\begin{center}
		\includegraphics[width=0.95\linewidth,height=4.5cm]{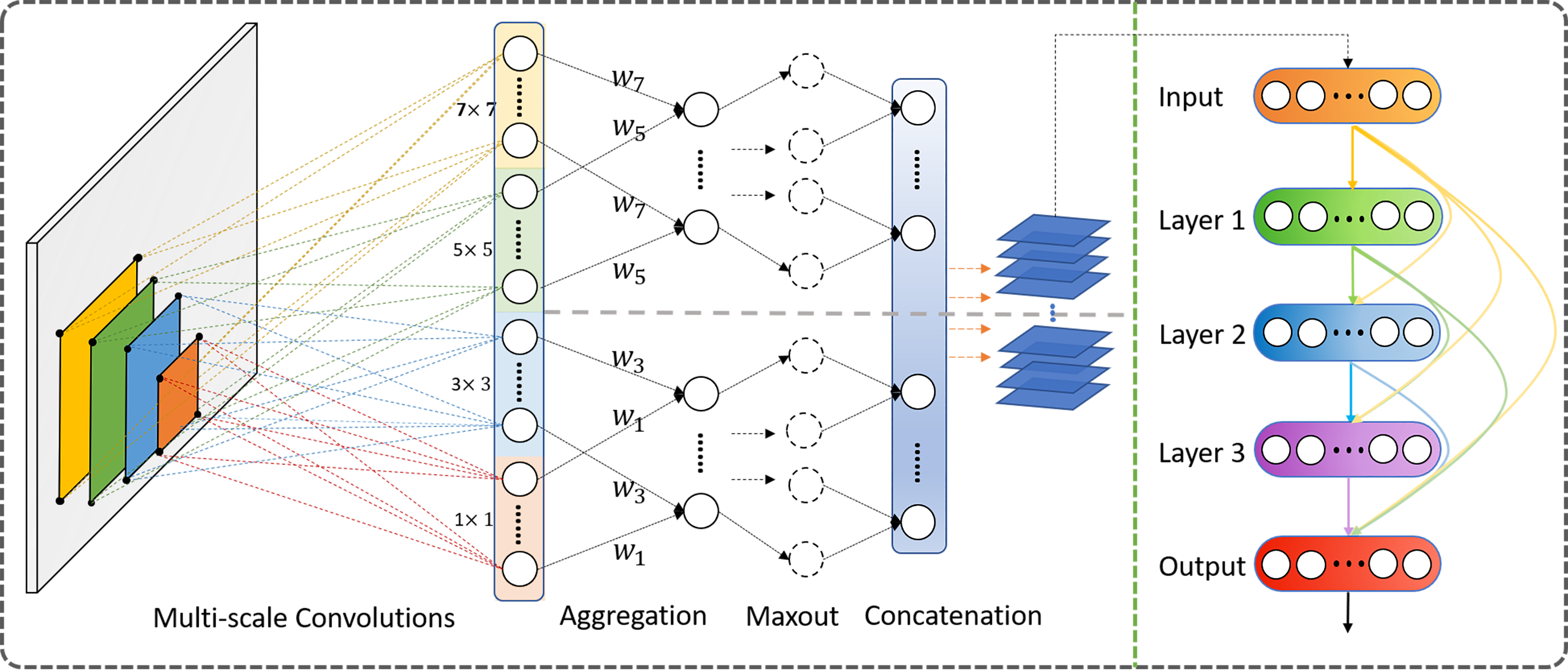}
	\end{center}
	\caption{DenseNets with Multi-scale Convolution Aggregation (MCA) module. Given the raw image on the left, the first layer generates four groups of feature maps using different kernel sizes. These results pass through aggregation and maxout gates to produce two branches of compressed channels, which capture fine and coarse scale features, respectively. The two channels are concatenated into a layer of feature maps, which is fed into {the DenseNets represented by with 3 composite layers on the right.}}
	\label{fig:1-1}
\end{figure}

Inspired by the benefits of multi-scale convolutions \cite{lin2017feature,yu2017deep} and features fusion for training deep networks, we design a novel module, referred as Multi-scale Convolution Aggregation (MCA) to work with DenseNets.  As shown in Fig.~\ref{fig:1-1}, the MCA module consists of layers for multi-scale convolutions, cross-scale aggregation, maxout, and concatenation. We observe that DenseNets utilizing MCA module can substantially reduce parameters number and classification error than using other multi-scale designs. The reduction in parameters results from the new design of fusing pyramidal convolutions instead of {simply concatenating them. The increase in accuracy is attributed to the following} factors: 1) strengthening scale-invariance because of the multi-scale convolutions with {four kernels with different receptive field sizes; 2) given a specific task, the network automatically chooses the most suitable scales} via four trainable gating units to adaptively make use of multi-scale information; 3) the use of two maxout activations {stimulates} the competition among neural units of different receptive fields and enhances the learning ability of the network; 4) higher non-linearity; and 5) compared with traditional concatenation in GoogleNets, our module dramatically reduces the number of parameters while preserving sufficient multi-scale information by aggregation and maxout functions. 

In addition to various methods of architecture design, difficulties in training deep networks motivated research on optimization and initialization techniques. These include dropout \cite{hinton2012improving}, maxout activation \cite{goodfellow2013maxout}, batch normalization \cite{ioffe2017batch,ioffe2015batch}, group normalization \cite{wu2018group}, Xavier initialization \cite{glorot2010understanding}, He initialization \cite{he2015delving}, etc., which have been applied in a wide range of networks as essential components.

To reduce the possibility of overfitting in DenseNets and to further boost the generalization of networks, we also develop a regularization method named Stochastic Feature Reuse (SFR). Similar to stochastic depth \cite{huang2016deep}, SFR contains gates for dropping selected feature maps delivered from preceding layers; see Fig.~\ref{fig:SFRstructure}. During training step, each layer randomly reuses different preceding feature maps for different mini-batch, resulting each mini-batch is trained under a sub-network with a unique connection scheme.  This approach effectively addresses overfitting problem of DenseNet by substantially reducing the number of parameters while improving the performance of DenseNets.

We evaluate the impacts of both MCA module and SFR on three widely used benchmark datasets: CIFAR-10 \cite{krizhevsky2009learning}, CIFAR-100 \cite{krizhevsky2009learning} and Street View House Number (SVHN) \cite{netzer2011reading}.  The comparisons show that our model can achieve comparable test accuracy with relatively lower computation costs and outperform the state-of-the-art performance of DenseNets.

\section{Related Work}

Deeper feed-forward neural networks tend to generate larger dividends in performances of various vision tasks.  {This leads to the recent resurgence of exploration in sophisticated CNNs architectures \cite{huang2017densely} with hugely increased classification accuracy on ImageNet \cite{deng2009imagenet}, e.g.} from AlexNet \cite{krizhevsky2012imagenet} to GoogLeNets \cite{szegedy2015going}, and ResNets \cite{he2016deep} to DenseNets \cite{huang2017densely}.

Comparisons of layerwise performance, analysis \cite{long2015fully,yosinski2014transferable} and visualization of feature maps \cite{yu2017dilated,zeiler2014visualizing} show that {networks with} deeper layers are able to extract more semantic and higher-level representations. {On the other hand, very deep networks make} training more difficult, especially when using a first-order optimizer with purely random initialization and traditional activation functions (tanh, sigmoid etc.), which {often} cause gradients vanishing and internal covariate shifts. To overcome these problems, a lot of research has been carried out \cite{larsson2016fractalnet}.


To deeply dig into high-performance architectures, a series of {independent} methods have been explored. One of more dominative is to increase the network width. GoogLeNets \cite{ioffe2015batch, simonyan2014very,szegedy2017inception,szegedy2016rethinking} use the inception module to build deep networks and this component concatenates feature maps produced by a set of filters with different receptive field size. Other well-known structures, such as Resnet in Resnet \cite{targ2016resnet} and Wide residual networks \cite{zagoruyko2016wide}, also demonstrate that simply increasing the number of filters in each layer can dramatically improve test accuracy. More recently, FractalNets \cite{larsson2016fractalnet} obtained excellent results using a wider block structure. In addition to increasing depth and width of networks, there are a growing number of research works focusing on aggregation or fusion. Deep Layer Aggregation \cite{yu2017deep} provides a novel approach to fuse features vertically across layers, which substantially improves recognition accuracy {with less computational cost}. 

{Inspired by these findings, we design a novel {MCA} module, which first broadens the width of the initial convolution layer of DenseNets through multi-scale convolutions, then fuses the filters using cross-scale aggregation parameterised by trainable weights.} The idea of multi-scale convolutions also follows a neuroscience model \cite{serre2007robust} suggesting that the raw image should be processed at different scales and then joined together for next layers, so that the deeper layers can become robust to scale shift \cite{szegedy2015going}.

{Another breakthrough in deep learning is the introduction of skip connections, which addresses the challenges of training deep networks.} Highway Networks \cite{srivastava2015highway} efficiently train deep networks by introducing the bypassing path, which is the primary factor that eases the training pain. ResNets \cite{he2016deep} further {enhance} this new connection pattern through substituting bypassing paths with residual connections, and achieve record-breaking performance on ImageNet \cite{deng2009imagenet}. Recently, DenseNets \cite{huang2017densely} densely connect all preceding layers with each layer to reuse all preceding feature maps and outperform the state-of-the-art results on several competitive benchmarks. Moreover, stochastic depth \cite{huang2016deep} was proposed as a successful approach to train an over 1000-layer ResNet through randomly dropping a few layers during training. Analogous to dropout \cite{hinton2012improving}, this method demonstrates that stochastically dropping is an extremely powerful technique to regularize networks. Our {SFR regularizer} was motivated by the observations on Dropout, Stochastic Depth and DenseNets. {However, instead of dropping layers as in Stochastic Depth, our regularizer drops features by randomly blocking a set of bypassing paths.}

\section{Methodology}

\textbf{DenseNets.} {Both the {MCA} module and the {SFR} regularizer proposed in this paper are} based on DenseNets \cite{huang2017densely}. Assume that a single input image is represented by $x$ and is passed through a DenseNet that has $L$ layers. Each layer $l$ comprises a composite function $H_l(\cdot)$ that includes one Batch Normalization layer \cite{ioffe2015batch}, one ReLU layer \cite{glorot2011deep}, and one $3\times3$ convolution layer. DenseNets introduce a new connectivity scheme: the output of each layer is directly connected to all subsequent layers. Consequently, the $l^{th}$ layer receives the outputs of all preceding layers. That is:
\begin{equation}
	x_l=H_l\left([x_0,x_1,...,x_{l-1}]\right)~.
	\label{equ:densenet}
\end{equation}
where $x_{l}$ is the output of layer $l$, $[x_0,x_1,...,x_{l-1}]$ is the concatenation of feature maps produced by layers 0, 1, 2, ..., $l-1$.  The total number of channels in a $L$-layer DenseNet, $N(L)$, can be approximatively computed as:
\begin{equation}
	N(L) \approx N(0)\times L+\frac{L(L-1)}{2}\times k~.   
	\label{equ:densenet}
\end{equation}
where $N(0)$ represents the number of input channels into first dense block and {$k$} is the growth rate of the DenseNet.

\subsection{Multi-scale Convolution Aggregation Module}

Through concatenating different groups of convolutions, Inception \cite{simonyan2014very} module and its variant \cite{liao2015competitive} have shown that multi-scale convolution filters can boost the performance of deep networks. Inspired by their findings, we design a novel {MCA} module to enhance the representative and learning capacity of DenseNets. {The new module} consists of layers for multi-scale convolutions, cross-scale aggregation, maxout, and concatenation.  It is placed in front of the DenseNet as initial layers so that abundant features extracted at different scales can be fed into the network.

\textbf{Multi-scale Convolutions.} Given the input image $x$, the multi-scale convolutions layer {computes the following}:
{\small
	\begin{equation}
		\begin{split}
			M_1(x,W)=G_{1\times1}(x,W_1)\circ G_{3\times3}(x,W_3) \\ 
			\circ G_{5\times5}(x,W_5) \circ G_{7\times7}(x,W_7)~,
			\label{equ:3.2.1}
		\end{split}
\end{equation}}
where $G_{n\times n}$ $(n=1,3,5,7)$ are the results of convolutions with $1\times1$, $3\times3$, $5\times5$, and $7\times7$ kernels respectively. $W_n$ represents parameters of different kernels and $``\circ"$ denotes the concatenation operator. 

Feeding the concatenation of four groups of convolutions, $M_1(x,W)$, into DenseNets directly helps to improve the performance of the network {since the network bandwidth is increased}.  When evaluated on CIFAR-10 dataset, a standard DenseNet with depth $L=40$ and growth rate $k=24$ achieves $93.45\%$, whereas the DenseNet with $M_1(x,W)$ as input achieves a test accuracy of 94.31$\%$. However, since and the number of initial channels for DenseNet, $N(0)$ in Equ.~(\ref{equ:densenet}), equals to the length of $M_1(x,W)$, {the number of parameters is increased} from 4.2 millions to 5.7 millions.

\begin{figure*}[t]
	\begin{center}
		\includegraphics[width=0.7\linewidth]{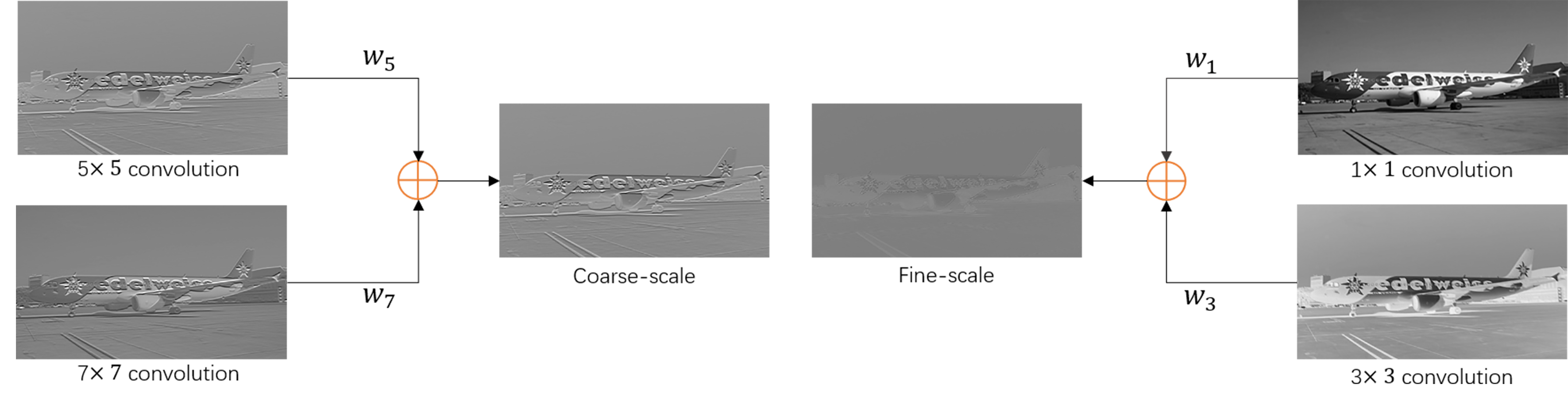}
	\end{center}
	\caption{{Illustration on} fine-scale and coarse-scale aggregations.}
	\label{fig:Aggregation}
\end{figure*}

\begin{figure}[t]
	\begin{center}
		\includegraphics[width=1.0\linewidth]{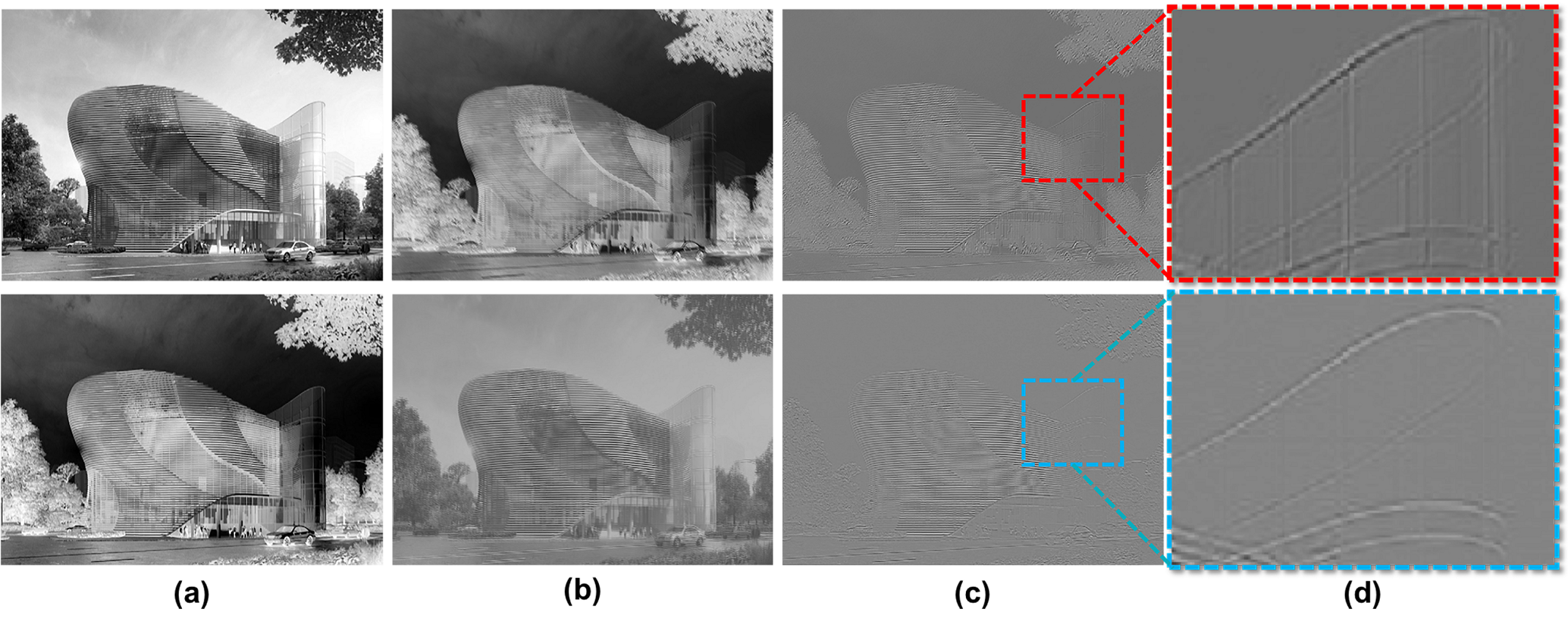}
	\end{center}
	\caption{Visualization on results obtained using the proposed adaptive aggregation (top row) and simple equal-weighted aggregation (bottom row): (a) results of $1\times 1$ convolutions; (b) results of $3\times 3$ convolutions; (c) aggregation of $1\times 1$ and $3\times 3$ convolutions; and (d) zoomed in view of dashed-box areas in (c), which shows that trainable aggregation can extract more abundant and detailed texture features.}
	\label{fig:AggregationVisual}
\end{figure}

\textbf{Cross-scale Aggregation.}  In order to reduce the complexity of the model while maintaining a high test accuracy and maximizing effective information flow of the network, an adaptive aggregation function is applied. Here we aggregate {convolution results under four {kernels} into two branches that represent fine and coarse scales, respectively; see Fig.~\ref{fig:Aggregation}. Since trainable gating weights are introduced, the unit is similar to a small-scale voting system. For each mini-batch, proper weights are automatically assigned to different scales via our voting mechanism. This helps to preserve most contributive multi-scale information and suppress flows with lower importance. Specifically,} the cross-scale aggregation layer performs the following operation: 
{\small
	\begin{equation}
		\begin{split}
			M_2(x,W)=&\left(w_1 G_{1\times1}(x,W_1) + w_3 G_{3\times 3}(x,W_3)\right) \\ 
			\circ &\left(w_5 G_{5\times 5}(x,W_5) + w_7 G_{7\times 7}(x,W_7)\right)~,
			\label{equ:3.2.2}
		\end{split}
\end{equation}}
where $``+"$ represents pixelwise summation aggregation. $w_1, w_3, w_5, w_7$ are learnable gating weights for {convolution results} at different scales. Their values indicate the importance of respective scales. In practice, we also found that trainable aggregation works much better than equal-weighted fusion, since {the voting system in the former approach} makes the module more adaptive on a variety of datasets. As shown in Fig.~\ref{fig:AggregationWeights}, the finally converged weights on different datasets vary widely, which {indicates that different datasets favor the contributions from different scales}. Fig.~\ref{fig:AggregationVisual} visually compares the results obtained through the two versions of the aggregation.

{Compared to the Inception module that simply concatenates different groups of convolutions, the aggregation layer we used can significantly reduces the number of parameters. On CIFAR-10 dataset, the number of parameters is reduced} from 5.7 millions to 4.2 millions in DenseNet with depth = 40 and growth rate $k = 24$. 

\textbf{Maxout.} {Previous work have shown} that: 1) maxout exploits the model averaging behavior as the approximation is more accurate; 2) back-forward flow of maxout can avoid pitfalls such as failing to use a large set of filters {\cite{goodfellow2013maxout}; and 3) grouping is important in deep networks \cite{wu2018group}.} Hence, to better regularize our fusion results, here two maxout operations are {independently performed after cross-scale aggregation layer, one for the two fines scale channels and the other for the two coarser scale channels.}
That is, we have the final output of MCA module $MC(x,W)$:
{\small
	\begin{equation}
		\begin{split}
			& Maxout\left(w_1G_{1\times1}(x,W_1)+w_3G_{3\times3}(x,W_3)\right)\\
			& \circ Maxout\left(w_5G_{5\times5}(x,W_5)+w_7G_{7\times7}(x,W_7)\right)~.   
			\label{equ:aggregation_1}
		\end{split}
	\end{equation}
}

With maxout layer introduced, the {whole MCA} module can be viewed as a highly non-linear transformation between original input and the first dense block of DenseNets. It includes four gating units parametrized by $w_i$ $(i=1,3,5,7)$ controlling the flow of multi-scale information.

\textbf{Backward Propagation.} The process of gradients back-propagation is the same as the traditional back-propagation. Here, we present the derivation formula in terms of weights of multi-scale convolutions; see Equ.~(\ref{equ:bp}). $x_1, x_3, x_5, x_7$ represent the outputs of multi-scale convolutions and $W_1, W_3, W_5, W_7$ are kernel weights. We define the maxout function as $M(x)$ and $H_{M}(x)$ {denotes} its first-order derivative. The input image is $x_0$ and $b_1, b_3, b_5, b_7$ are bias vectors.  
{\small
	\begin{equation}
		\begin{split}
			&x_1=W_1x_0+b_1,x_3=W_3x_0+b_3\\
			&x_5=W_5x_0+b_5,x_7=W_7x_0+b_7\\
			&x=M(w_1(x_1)+w_3(x_3)) \circ M(w_5(x_5)+w_7(x_7))\\
			&\frac{\alpha \mathcal{L}}{\alpha b^1}=\delta^1,\frac{\alpha \mathcal{L}}{\alpha x}=(W^1)^T\delta^1=\delta^0\\
			&\frac{\alpha \mathcal{L}}{\alpha x_1}=\frac{\alpha \mathcal{L}}{\alpha x}\frac{\alpha x}{\alpha x_1}=H_{M}(x)w_1\delta^0,\frac{\alpha \mathcal{L}}{\alpha x_3}=H_{M}(x)w_3\delta^0\\
			&\frac{\alpha \mathcal{L}}{\alpha x_5}=H_{M}(x)w_5\delta^0,\frac{\alpha \mathcal{L}}{\alpha x_7}=H_{M}(x)w_7\delta^0\\
			&\frac{\alpha \mathcal{L}}{\alpha W_1}=\frac{\alpha \mathcal{L}}{\alpha x_1}\frac{\alpha x_1}{\alpha W_1}=H_{M}(x)w_1\delta^0x_0\\
			&\frac{\alpha \mathcal{L}}{\alpha W_3}=\frac{\alpha \mathcal{L}}{\alpha x_3}\frac{\alpha x_3}{\alpha W_3}=H_{M}(x)w_3\delta^0x_0\\
			&\frac{\alpha \mathcal{L}}{\alpha W_5}=\frac{\alpha \mathcal{L}}{\alpha x_5}\frac{\alpha x_5}{\alpha W_5}=H_{M}(x)w_5\delta^0x_0\\
			&\frac{\alpha \mathcal{L}}{\alpha W_7}=\frac{\alpha \mathcal{L}}{\alpha x_7}\frac{\alpha x_7}{\alpha W_7}=H_{M}(x)w_7\delta^0x_0~.
			\label{equ:bp}
		\end{split}
\end{equation}}
where $\mathcal{L}$ is the loss function of the whole network and $W^1, b^1$ are the weight and bias of the first layer in the first dense block. $\delta^l$ is the sensitivity of $l^{th}$ layer. 

\subsection{Stochastic Feature Reuse}

Dropout \cite{hinton2012improving}, Drop-connect \cite{wan2013regularization} and Maxout \cite{goodfellow2013maxout} provide excellent regularization methods through modifying interactions among neural units or connections between different layers in order to break co-adaptation. These techniques have been supported by subsequent research and applied in a wide range of network architectures, such as ResNets \cite{he2016deep} and FractalNets \cite{larsson2016fractalnet}. Recent stochastic depth \cite{huang2016deep} and drop-path \cite{larsson2016fractalnet} successfully extend dropout and make impressive progress in vision tasks.

\begin{figure*}
	\begin{center}
		\includegraphics[width=0.9\linewidth]{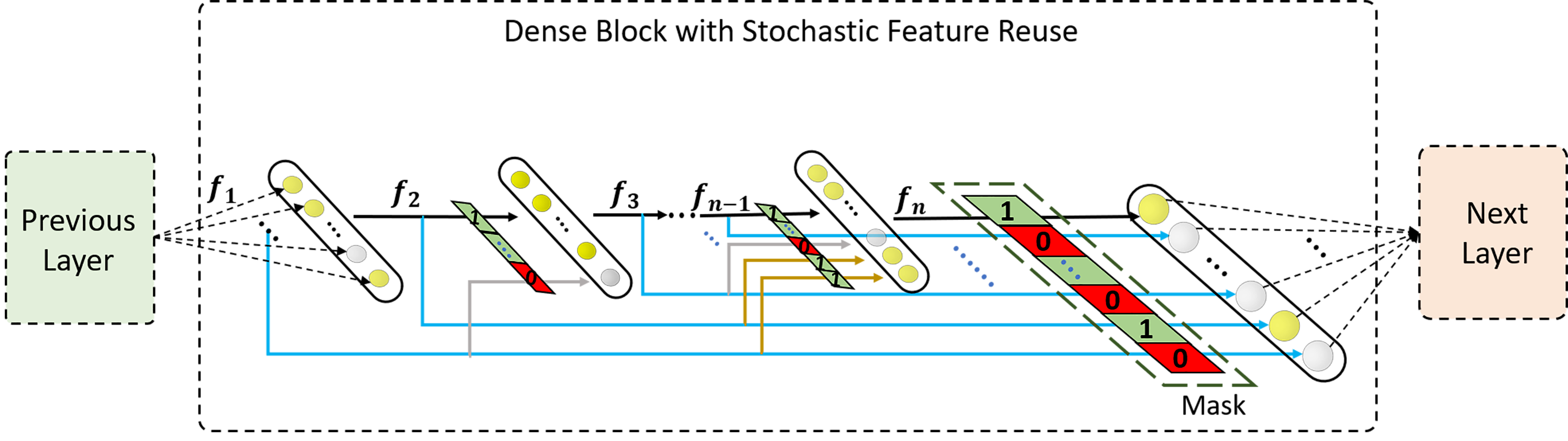}
	\end{center}
	\caption{At a given layer of a dense block, the original DenseNets concatenate of all feature maps produced by preceding layers as input. The presented Stochastic Feature Reuse uses a mask to drop features from some of these layers but guarantees at least one set of feature maps from previous layers will be reused. The mask is randomly generated and changes for different mini-batches during training.}
	\label{fig:SFRstructure}
\end{figure*}

Motivated by these structures, we propose ``Stochastic Feature Reuse'' (SFR) as an effective regularizer in DenseNets to promote the generalization of networks and to overcome overfitting especially when the growth rate is high.  Fig.~\ref{fig:SFRstructure} illustrates the model of SFR. For each mini-batch, a new mask tensor $M_l$ obeying \emph{Bernoulli distribution} is randomly generated for each layer $l$ and the input of $l+1$ layer is modified as follows:
\begin{equation}
	I_{l+1}=M_lx_l=M_l\cdot H_l\left([x_0,x_1,...,x_{l-1}]\right)~.
	\label{equ:drop}
\end{equation}
During the training time, when a set of skip connections are blocked, there is no need to perform forward and backward computations trough those. Hence, these dropped features are not reused by the current layer. Since a large amount of computation is saved, SFR can speed up the convergence of the network. When testing, all features are reused in order to make use of the full-width network \cite{huang2016deep}. 

As a regularizer, SFR can enhance the performance of DenseNets and deal with {the overfitting issue \cite{huang2017densely} through discouraging co-adaptation.  In addition, SFR also implicitly trains an ensemble of DenseNets, which helps to improve the performance.} For a $L$-layer DenseNet, there are $2^{\frac{L(L+1)}{2}}$ possible combinations and the final network used at the testing stage can be viewed as the average of these sub-networks.

\section{Experiments}

The presented {MCA module and SFR regularizer} are evaluated using three widely adopted benchmarks:  CIFAR-10 \cite{krizhevsky2009learning}, CIFAR-100 \cite{krizhevsky2009learning} and SVHN \cite{netzer2011reading}. The results show that {the performance of DenseNets with MCA modules is superior to the original DenseNets and that the SFR regularizer can effectively prevent overfitting}.

\subsection{Implementation and Training Details}
In our experiments, we report test error from the epoch with the lowest validation error and we use the same construction and training scheme as introduced {in DenseNet} \cite{huang2017densely}. {When evaluating the MCA module, the DenseNet part has three dense blocks, all have equal numbers of layers and the same growth rate. When evaluating SFR regularizer, an additional dense block with SFR is added so that the performance of the original DenseNet is not affected.} Each composite function of dense block uses a $3\times3$ convolution layer with zero-padding to keep the feature maps fixed. Between two dense blocks, there are bottleneck layers with a compression factor. In this paper, we set compression factor as 1.0 in standard DenseNet while set as 0.5 in the structure of DenseNet with bottleneck and compression (DenseNet-BC). At the end of the last dense block, a global average pooling layer, followed by a softmax layer, is attached. The sizes of feature maps in each of the three dense blocks are $32\times32$, $16\times16$ and $8\times8$, respectively.

Similar to the standard DenseNet \cite{huang2017densely}, DenseNets in our experiments are optimized through the first-order SGD optimizer. We train 350 epochs for CIFAR and 40 epochs for SVHN. Initial learning rate is 0.1 and divided by 10 at epochs 150, 225 and 300 for CIFAR and epochs 20 and 30 for SVHN. We also add weight decay (0.0001) term into our loss function and use Nesterov momentum \cite{sutskever2013importance} of 0.9 for optimization. Hinton's Dropout \cite{hinton2012improving} layer with drop probability $p=0.2$, Batch Normalization \cite{ioffe2015batch} layer and He Initialization of weights \cite{he2015delving} are applied as well.

\begin{table*}[t]
	\begin{center}
		\begin{tabular}{|c|c|c|c|c|c|}
			\hline
			Model$\qquad\qquad$& Depth & Params. & C10($\%$)& C100($\%$) & SVHN($\%$)\\
			\hline
			Stochastic Pooling \cite{zeiler2013stochastic} & - & - & 15.13 & 42.51 & 2.80\\
			\hline
			Maxout Networks \cite{goodfellow2013maxout} &  - & - & 11.68 & 38.57 & 2.47\\
			\hline
			Network in Network \cite{lin2013network}  &  - & - & 10.41 & 35.68 & 2.35\\
			\hline
			Deeply Supervised Net \cite{lee2015deeply}  &  - & - & 9.69 & $34.57^+$ & 1.92\\
			\hline
			Competitive Multi-scale \cite{liao2015competitive}  &  - & 4.48M & 6.87 & 27.56 & 1.76\\
			\hline
			Highway Network \cite{srivastava2015highway}  &  - & - & $7.72^+$ & $32.39^+$ & -\\
			\hline
			Fractal Network \cite{larsson2016fractalnet}  &  21 & 38.6M & 10.18 & 35.34 & 2.01\\
			\hline
			FractalNet with Drop-path \cite{larsson2016fractalnet}  &  21 & 38.6M & 7.33 & 28.20 & 1.87\\
			\hline
			ResNet \cite{he2016deep}  &  110 & 1.7M & $6.61^+$ & - & -\\
			\hline
			Stochastic Depth \cite{huang2016deep}  &  110 & 1.7M &11.66 & 37.80 & 1.75\\
			\hline
			ResNet(pre-activation) \cite{he2016identity}  &  164 & 1.7M & 11.26 & 35.58 & -\\
			&  1001 & 10.2M & 10.56 & 33.47 & -\\
			\hline
			DenseNet($k=12$) \cite{huang2017densely}  &  40 & 1.0M & 7.00 & 27.55 & 1.79\\
			\hline
			DenseNet($k=24$) \cite{huang2017densely}  &  100 & 27.2M & 5.83 & 23.42 & 1.59\\
			\hline
			DenseNet($k=24$)\cite{huang2017densely}  &  53 & 7.8M & 6.45 & 24.32 & 1.78\\
			\hline
			DenseNet with SFR($k=24$)  &  53 & 7.8M & {\bf 6.08} & {\bf 23.82} & {\bf 1.66}\\
			\hline
			DenseNet-BC($k=12$)\cite{huang2017densely}  &  100 & 0.8M & 5.92 & 24.15 & 1.76\\
			\hline
			DenseNet-BC with MCA($k=12$)  &  100 & 0.8M & {\bf 5.41} & {\bf  24.07} & {-}\\
			\hline
			DenseNet with MCA($k=12$)   & 40 & 1.0M & { 6.44} & { 27.44} &  { 1.77}\\
			\hline
			DenseNet with MCA($k=24$)   & 40 & 4.2M & {$\bf 5.38^{*}$} & {\bf 23.78} &  {\bf 1.66}\\
			\hline
			DenseNet with MCA($k=40$)   & 40 & 11.6M & { 5.76} & {$\bf 22.65^*$} &  {$\bf 1.61^*$}\\  
			\hline
		\end{tabular}
	\end{center}
	\caption{
		Test error on CIFAR and SVHN datasets. Contents in boldface are our competitive results. $``+"$ indicates that the error rate is based on datasets with data augmentations. {DenseNet with MCA achieves better performance than the original under the same configuration. Particularly, when growth rate, depth are set as 24 and 40, the network obtains an excellent result (${\bf 5.38\%}$) on CIFAR-10 which is better than the original DenseNet with 100 and 53 layers. On CIFAR-100 and SVHN, our model with $k=40$ achieves more remarkable results (${\bf 22.65\%}$ and ${\bf 1.61\%}$). In the structure of DenseNet-BC, our MCA also has positive impacts on the performance.} } 
	\label{table:drop}
\end{table*}

\subsection{Datasets}

The {\bf CIFAR-10} dataset \cite{krizhevsky2009learning} consists of 60,000 {(50,000 for training + 10,000 for testing)} {natural color images of 32$\times$32 resolution}. {Objects from ten classes (e.g. vehicles, flowers etc.) have equal volume of training and test images and are centered in these images}. The {\bf CIFAR-100} dataset extends the number of classes in CIFAR-10 to 100, but each class only consists of 600 images. Due to more classes and fewer samples for each class, the classification for CIFAR-100 {is} considered as more challenging. {\bf Street View House Number (SVHN)} dataset is also a well-known benchmark in {computer vision, which consists of color images of digits 0 to 9 of 32$\times$32 resolution.} There are 73,257 training, 26,032 testing, and 531,131 additional training images respectively. 

In our experiments, we apply the same normalization methods on input images as the original DenseNet. For CIFAR dataset, we subtract mean values and divide standard deviations{, whereas for SVHN images, the pixel values were divided by 255.} We do not use any data augmentation in the experiments, and only focus on comparing our approaches with other {network} models on original datasets.

\subsection{Results and Discussion}
We train our networks with different depths (40, 53, 100) and growth rates $k$ ($k=12,24,40$) and compare our approach with other well-known models on CIFAR-10, CIFAR-100, SVHN; see Table 1.

\begin{figure*}[t]
	\begin{center}
		\includegraphics[width=0.80\linewidth,height=3.9cm]{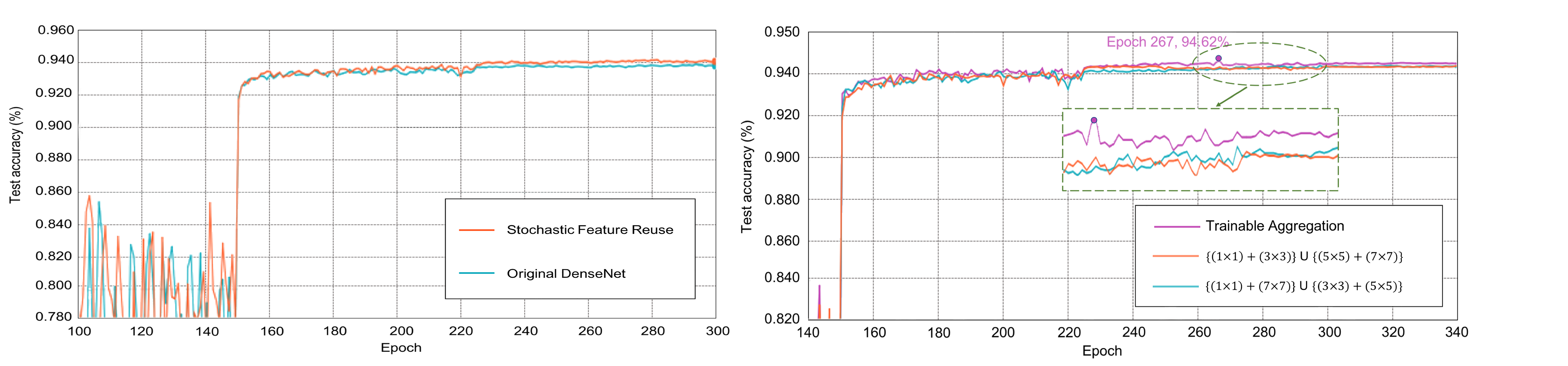}
	\end{center}
	\caption{Test accuracy on CIFAR-10. {All structures consist of the MCA module. {\em Left:} comparison between DenseNets with and without SFR regularizer. Our dropout has a constant drop rate of $20\%$. {\em Right:} comparison of three aggregation patterns, which shows that adaptive fusion is more powerful and representative. }} 
	\label{fig:difaggregation}
\end{figure*}

\paragraph{Multi-scale Convolution Aggregation.} To better evaluate our novel module, we train different patterns of aggregation on CIFAR-10 and test the best model on CIFAR-10, CIFAR-100 and SVHN. The performance of our structure with different setting on three benchmarks are shown in the bottom of Table 1. With relatively fewer parameters (4.2M), it obtains the lowest classification error rate on CIFAR-10 (5.38\%)  and CIFAR-100 (23.78\%), and second best results on SVHN (1.66\%). {In the case of $k=40$, depth = 40, our model gets { impressive} results (22.65\%) on CIFAR-100 and (1.61\%) on SVHN.} This demonstrates that our MCA module has much higher representative capacity and is able to preserves abundant information of multi-scale convolutions. This is crucial for preventing overfitting and promoting generalization ability.  

{Fig.~\ref{fig:difaggregation}(right) compares different aggregation patterns for fussing multi-scale convolutions information ($k=24$, depth = 40). The aggregation parameterized} by gating weights gains the best performance with only four parameters added. Its success may be attributed to the following factors:

{
	\textbf{Factor 1:} Aggregating different scales with trainable weights is more flexible and representative than aggregating} with hand-crafted weights. During the process of SGD, the weights of different kernel sizes are treated {independently and adaptively. Since pixels at different distances from the central point should have different importance, this strategy can preserve richer multi-scale information (texture, edges, corners, etc.) while using much fewer parameters than simply concatenating them}.

\begin{figure*}[t]
	\begin{center}
		\includegraphics[width=0.75\linewidth,height=7.5cm]{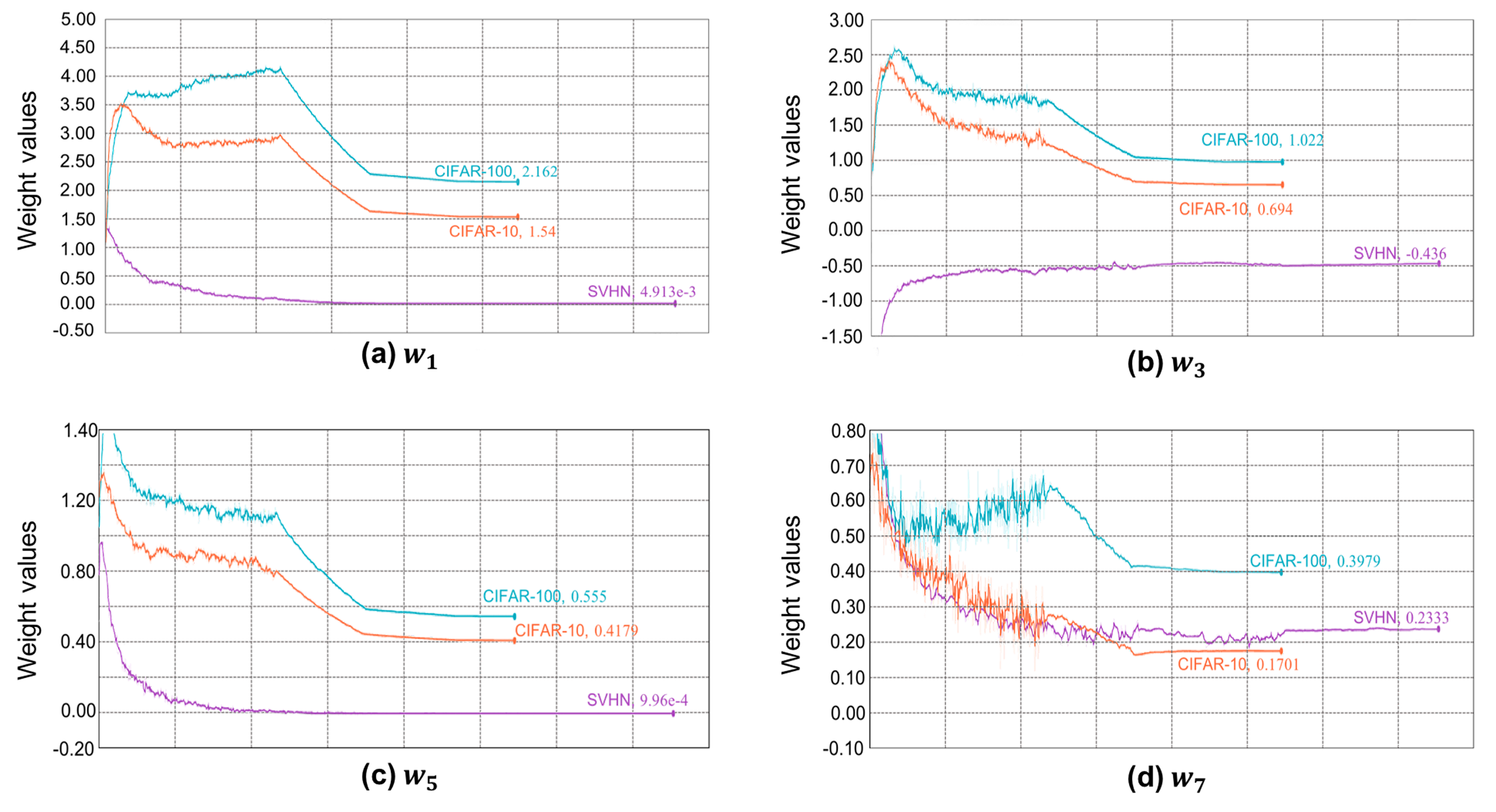}
	\end{center}
	\caption{{The variation of aggregation weights during the training under different datasets. The optimal weights for different tasks are different.} Under the {Stochastic Gradient Descend (SGD)} , the model adaptively controls the flow of multi-scale information {so that the scales with high discrimination power are preserved whereas the redundant ones are suppressed.}} 
	\label{fig:AggregationWeights}
\end{figure*}

More importantly, for vision tasks with different complexity, {weights of gating units may vary under similar trends during training, but often converge to different final values; see Figure~\ref{fig:AggregationWeights}.
	This suggests that the optimal scale for convolutions can be different for different datasets.} For instance, in CIFAR tasks, the module {assigns high weights ($w_1$ and $w_3$) to fine-scale features, whereas less coarse-scale information is delivered to subsequent DenseNet. 
	On the other hand, for the SVHN dataset, the weight $w_7$ has much higher relative value than for the other two dataset, whereas the weight $w_1$ is almost 0. {This observation suggests that,} for simple digits classification tasks, coarse-scale features extracted by $7\times 7$ convolution is more important than in other more complicated tasks.} To further demonstrate this point, we also run our module on another simple dataset MNIST and obtain the similar observation {($w_7=0.5456$ for MNIST vs. $w_7=0.1706$ for CIFAR-10)}. 

\textbf{Factor 2:} The combination of three dominant joining methods (summation, maxout and concatenation) makes our model highly non-linear and capable of effectively aggregating multi-scale representations. Each joining method has its own advantages. The combination of different approaches is also studied in \cite{szegedy2017inception}, which shows a better performance. By utilizing two maxout, the units of the aggregation layer have strong competition which is beneficial for training and optimizing deep networks. The two branches of the fine-scale and coarse-scale aggregations enhance the scale invariance property.

\paragraph{Stochastic Feature Reuse.} We evaluate SFR on the same three datasets {and compare it} with the original DensNet with depth $= 53$ and growth rate $k=24$. {The additional dense block with SFR is placed at the front or at the end of the original DenseNet; see {Table \ref{table:dropon4th}} for details.  The comparison shows that placing the additional dense block with SFR at the end of the DenseNet generates lower error rates on all three datasets.} We attribute the accuracy improvement to the fact that SFR randomly generates a new sub-network with different propagation path for each mini-batch and implicitly train an ensemble of different networks. This kind of dropout can disorganize the co-adaptation among reused features and prevent overfitting. {On the other hand, adding the additional dense block with SFR to the front of DenseNet actually hurt the performance since this will lead that shallow layers are too narrow to pass sufficient information flow.} In addition, we observe that SFR should work with Hinton's Dropout, without which the accuracy also degenerates.

\begin{table}
	{\small
		\begin{center}
			\begin{tabular}{|c|c|c|c|}
				\hline
				$\qquad\qquad$& Block index & Error($\%$) & Dataset\\
				\hline
				DenseNet \cite{huang2017densely} & $None$ & 6.45&{\small CIFAR-10} \\
				SFR & $4^{th}$ & {\bf 6.08} & {\small CIFAR-10}\\
				SFR & $1^{st}$ & 8.99 &{\small CIFAR-10} \\
				SFR(No Dropout) & $4^{st}$ & 10.00 &{\small CIFAR-10} \\
				\hline
				DenseNet \cite{huang2017densely} & None & 24.32 &{\small CIFAR-100} \\
				SFR & $4^{st}$ & {\bf 23.82} &{\small CIFAR-100}\\
				SFR & $1^{st}$ & 26.54 & {\small CIFAR-100} \\
				SFR(No Dropout) & $4^{st}$ & 27.15 &{\small CIFAR-100} \\
				\hline
				DenseNet \cite{huang2017densely} & None & 1.78 &  {\small SVHN}\\
				SFR & $4^{st}$ & {\bf 1.66} & {\small SVHN}\\
				SFR & $1^{st}$ & 2.12 & {\small SVHN}\\
				SFR(No Dropout) & $4^{st}$ & 3.02 &{\small SVHN}\\
				\hline
			\end{tabular}
		\end{center}
	}
	\caption{
		Test error of DenseNets trained with stochastic feature reuse on different datasets without data augmentation. SFR is more powerful when placed at the end of DenseNets.  
	}
	\label{table:dropon4th}
\end{table}

{Another observation} is that our method is more {effective} on {wider} DenseNets, as narrow networks in general do not have serious co-adaptation {issue or} long training time. 
{The ways of widening DenseNet mainly includes using a larger growth rate $k$ or increasing channels of the first initial layer. Hence, to illustrate the impact of different growth rate on the performance of our SFR, we firstly evaluate on CIFAR-10 based on three growth rates 12, 24 and 40. Moreover, the case of wider initial layer also be considered. Here we expand the first initial convolution layer four times  via four-scale convolutions and the training results are shown in Fig.~\ref{fig:difaggregation}(left). Table \ref{table:dropk} shows the results under different cases. SFR test error increases to 6.32\% when growth rate adds up to 40 as a very large bandwidth causes slight overfitting. Using {SRF with high drop probability addresses} this issue.} 


\begin{table}
	{\small
		\begin{center}
			\begin{tabular}{|c|c|c|c|c|}
				\hline
				$\qquad\qquad$ &Width& {w.o. SFR} &{w. SFR} & {Improve}\\
				\hline
				SFR($k=12$) & 17196 & 6.93 & 6.80 & 0.13\\
				\hline
				SFR($k=24$) & 34392 &6.45 & 6.08 & {\bf 0.37}\\
				\hline
				SFR(WIL) & 34536 & 6.09 & 5.76& 0.33\\	
				\hline
				SFR($k=40$) & 57320 &6.53 & 6.32 & 0.21\\	
				\hline	
			\end{tabular}
		\end{center}
	}
	\caption{Test error {with or without SFR under} different growth rates $k$ and wider initial layer on CIFAR-10. {WIL means ``Wider Initial Layer". Width is the maximal number of channels of all layers in DenseNet. When growth rate is set as 24, our SFR is more beneficial for improvements of performance on CIFAR-10.}   
	}
	\label{table:dropk}
\end{table}

\section{Conclusions}

A novel network module, referred as \emph{Multi-scale Convolution Aggregation}, is presented in this paper.  It consists of 4 groups of multi-scale convolutions, {cross-scale aggregation parametrized by 4 trainable weights and 2 maxout that produces 2 branches of feature maps representing smaller and larger receptive fields respectively.} In our experiments, Densenets with our new model obtain excellent performance while requiring substantially fewer parameters than utilizing traditional inception module. Instead of simple equal-weighted aggregation, our aggregation employs self-adaptive strategy to control the information flow of convolution filters. It automatically optimizes these weights according to different vision tasks. Trainable aggregation guarantees the maximum use of multi-scale convolutions and is the key {for} reducing parameters, {whereas} maxout strengthens the competitions among units in fine-scale and coarse-scale branches. The combination of three joining methods: concatenation, summation and maxout makes the networks {highly} non-linear.

In addition, a \emph{Stochastic Feature Reuse} strategy is also presented for training deep DenseNets effectively and efficiently. This regularizer downsamples a new subnets of the basic DenseNet for each mini-batch during training but reuses all feature maps produced by preceding layers at test stage.  Our method {enhances the performance of DenseNets through breaking the co-adaptation among reused features and implicitly training an ensemble of multi-subnets with different widths. Being a simple and easy-to-apply approach, SFR is more useful for {wider} DenseNets with a larger growth rate and can effectively alleviate the difficulties of training wide networks.

	For future work, we would like to explore the applications of the MCA module in other prominent deep architectures, as we felt MCA can be beneficial through introducing scale-invariance information without adding feature redundancy.  In addition, when evaluating SFR, we empirically use a constant drop probability.  It is interesting and meaningful to explore other configurations of the drop probability in future experiments.
}


{\small
\bibliographystyle{ieee}
\bibliography{egbib}
}

\end{document}